\documentclass{article} % For LaTeX2e
\usepackage{nips14submit_e,times}

\usepackage{graphicx}
\usepackage{xspace}
\usepackage{rotating}
\usepackage{multirow} 
\usepackage{cite}
\usepackage{url}
\usepackage{verbatim}

\usepackage{amsmath}
\usepackage{amssymb}
\usepackage{bbm} % blackboard math \mathbbm{1}, symbols-a4.pdf, p. 40
\usepackage{bm} % \bm{\alpha}

\usepackage{pstricks}
\usepackage{pst-plot}
\usepackage[pstadd]{pst-sigsys}
\usepackage{pstricks-add}
\usepackage{multido}

\usepackage[marginclue,inline,final]{fixme}

\newcommand{\norm}[1]{\ensuremath{\lVert#1\rVert}}

\newcommand{\lspan}[1]{\ensuremath{\mathrm{span}}}
\newcommand{\by}[2]{\ensuremath{#1 \! \times \! #2}}

\newcommand{\xs}{\ensuremath{x}}        % 

\newcommand{\vv}{\ensuremath{\mathbf{v}}}        % 
\newcommand{\wv}{\ensuremath{\mathbf{w}}}        % 
\newcommand{\xv}{\ensuremath{\mathbf{\xs}}}        % 
\newcommand{\fig}{Fig.}

\makeatletter 
\DeclareRobustCommand\onedot{\futurelet\@let@token\@onedot}
\def\@onedot{\ifx\@let@token.\else.\null\fi\xspace}

\def\etal{et al\onedot} 
\makeatother

\usepackage{hyperref}

\title{Deep Epitomic Convolutional Neural Networks}

\author{%
George Papandreou\\
Toyota Technological Institute at Chicago\\
{\tt\small gpapan@ttic.edu}
}

\nipsfinalcopy % Uncomment for camera-ready version

\begin{document}
\sloppy

\maketitle

\begin{abstract}
  Deep convolutional neural networks have recently proven extremely
  competitive in challenging image recognition tasks. This paper proposes the
  epitomic convolution as a new building block for deep neural networks. An
  epitomic convolution layer replaces a pair of consecutive convolution and
  max-pooling layers found in standard deep convolutional neural networks. The
  main version of the proposed model uses mini-epitomes in place of filters
  and computes responses invariant to small translations by epitomic search
  instead of max-pooling over image positions. The topographic version of the
  proposed model uses large epitomes to learn filter maps organized in
  translational topographies. We show that error back-propagation can
  successfully learn multiple epitomic layers in a supervised fashion. The
  effectiveness of the proposed method is assessed in image classification
  tasks on standard benchmarks. Our experiments on Imagenet indicate improved
  recognition performance compared to standard convolutional neural networks
  of similar architecture. Our models pre-trained on Imagenet perform
  excellently on Caltech-101. We also obtain competitive image classification
  results on the small-image MNIST and CIFAR-10 datasets.
\end{abstract}

\section{Introduction}
\label{sec:intro}

Deep learning offers a powerful framework for learning increasingly complex
representations for visual recognition tasks. The work of Krizhevsky \etal
\cite{KSH13} convincingly demonstrated that deep neural networks can be very
effective in classifying images in the challenging Imagenet benchmark
\cite{DDSL+09}, significantly outperforming computer vision systems built on
top of engineered features like SIFT \cite{Lowe04}. Their success spurred a
lot of interest in the machine learning and computer vision
communities. Subsequent work has improved our understanding and has refined
certain aspects of this class of models \cite{ZeFe13b}. A number of different
studies have further shown that the features learned by deep neural networks
are generic and can be successfully employed in a black-box fashion in other
datasets or tasks such as image detection \cite{ZeFe13b, OuWa13, SEZM+14,
  GDDM14, RASC14, CSVZ14}.

The deep learning models that so far have proven most successful in image
recognition tasks are feed-forward convolutional neural networks trained in a
supervised fashion to minimize a regularized training set classification error
by back-propagation. Their recent success is partly due to the availability of
large annotated datasets and fast GPU computing, and partly due to some
important methodological developments such as dropout regularization and
rectifier linear activations \cite{KSH13}. However, the key building blocks of
deep neural networks for images have been around for many years \cite{LBBH98}:
(1) convolutional multi-layer neural networks with small receptive fields that
spatially share parameters within each layer. (2) Gradual abstraction and
spatial resolution reduction after each convolutional layer as we ascend the
network hierarchy, most effectively via max-pooling \cite{RiPo99, JKRL09}.

In this work we build a deep neural network around the epitomic representation
\cite{JFK03}. The image epitome is a data structure appropriate for learning
translation-aware image representations, naturally disentagling appearance and
position modeling of visual patterns. In the context of deep learning, an
epitomic convolution layer substitutes a pair of consecutive convolution and
max-pooling layers typically used in deep convolutional neural networks. In
epitomic matching, for each regularly-spaced input data patch in the lower
layer we search across filters in the epitomic dictionary for the strongest
response. In max-pooling on the other hand, for each filter in the dictionary
we search within a window in the lower input data layer for the strongest
response. Epitomic matching is thus an input-centered dual alternative to the
filter-centered standard max-pooling.

We investigate two main deep epitomic network model variants. Our first
variant employs a dictionary of mini-epitomes at each network layer. Each
mini-epitome is only slightly larger than the corresponding input data patch,
just enough to accomodate for the desired extent of position invariance. For
each input data patch, the mini-epitome layer outputs a single value per
mini-epitome, which is the maximum response across all filters in the
mini-epitome. Our second topographic variant uses just a few large epitomes at
each network layer. For each input data patch, the topographic epitome layer
outputs multiple values per large epitome, which are the local maximum
responses at regularly spaced positions within each topography.

We quantitatively evaluate the proposed model primarily in image
classification experiments on the Imagenet ILSVRC-2012 large-scale image
classification task. We train the model by error back-propagation to minimize
the classification log-loss, similarly to \cite{KSH13}. Our best mini-epitomic
variant achieves 13.6\% top-5 error on the validation set, which is 0.6\%
better than a conventional max-pooled convolutional network of comparable
structure whose error rate is 14.2\%. Note that the error rate of the original
model in \cite{KSH13} is 18.2\%, using however a smaller network. All these
performance numbers refer to classification with a single network. We also
find that the proposed epitomic model converges faster, especially when the
filters in the dictionary are mean- and contrast-normalized, which is related
to \cite{ZeFe13b}. We have found this normalization to also accelerate
convergence of standard max-pooled networks. We further show that a deep
epitomic network trained on Imagenet can be effectively used as black-box
feature extractor for tasks such as Caltech-101 image classification. Finally,
we report excellent image classification results on the MNIST and CIFAR-10
benchmarks with smaller deep epitomic networks trained from scratch on these
small-image datasets.

\paragraph{Related work}

Our model builds on the epitomic image representation \cite{JFK03}, which was
initially geared towards image and video modeling tasks. Single-level
dictionaries of image epitomes learned in an unsupervised fashion for image
denoising have been explored in \cite{AhEl08, BMBP11}. Recently, single-level
mini-epitomes learned by a variant of K-means have been proposed as an
alternative to SIFT for image classification \cite{PCY14}. To our knowledge,
epitomes have not been studied before in conjunction with deep models or
learned to optimize a supervised objective.

The proposed epitomic model is closely related to maxout networks
\cite{GWMCB13}. Similarly to epitomic matching, the response of a maxout layer
is the maximum across filter responses. The critical difference is that the
epitomic layer is hard-wired to model position invariance, since filters
extracted from an epitome share values in their area of overlap. This
parameter sharing significantly reduces the number of free parameters that
need to be learned. Maxout is typically used in conjunction with max-pooling
\cite{GWMCB13}, while epitomes fully substitute for it. Moreover, maxout
requires random input perturbations with dropout during model training,
otherwise it is prone to creating inactive features. On the contrary, we
have found that learning deep epitomic networks does not require dropout in
the convolutional layers -- similarly to \cite{KSH13}, we only use dropout
regularization in the fully connected layers of our network.

Other variants of max pooling have been explored before. Stochastic pooling
\cite{ZeFe13a} has been proposed in conjunction with supervised
learning. Probabilistic pooling \cite{LGRN09} and deconvolutional networks
\cite{ZKTF10} have been proposed before in conjunction with unsupervised
learning, avoiding the theoretical and practical difficulties associated with
building probabilistic models on top of max-pooling. While we do not explore
it in this paper, we are also very interested in pursuing unsupervised
learning methods appropriate for the deep epitomic representation.

The topographic variant of the proposed epitomic model naturally learns
topographic feature maps. Adjacent filters in a single epitome share values in
their area of overlap, and thus constitute a hard-wired topographic
map. This relates the proposed model to topographic ICA \cite{HyHo01}  and
related models \cite{OWH06, KRFL09, LRMD+12}, which are typically trained to
optimize unsupervised objectives.

\section{Deep Epitomic Convolutional Networks}
\label{sec:model}

\begin{figure}[!tbp]
  \centering
  \begin{tabular}{c c}
    \includegraphics[width=0.45\columnwidth]{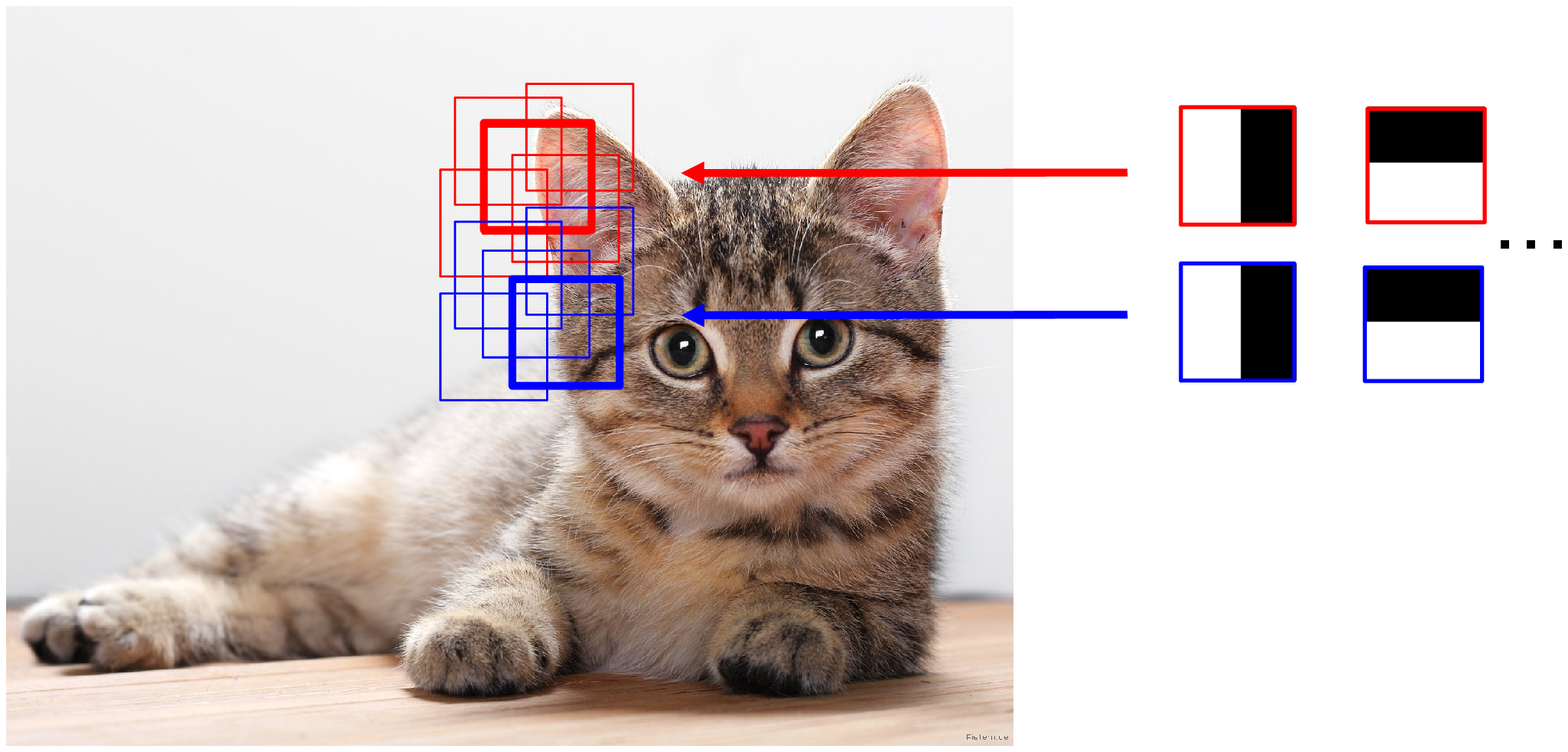}&
    \includegraphics[width=0.45\columnwidth]{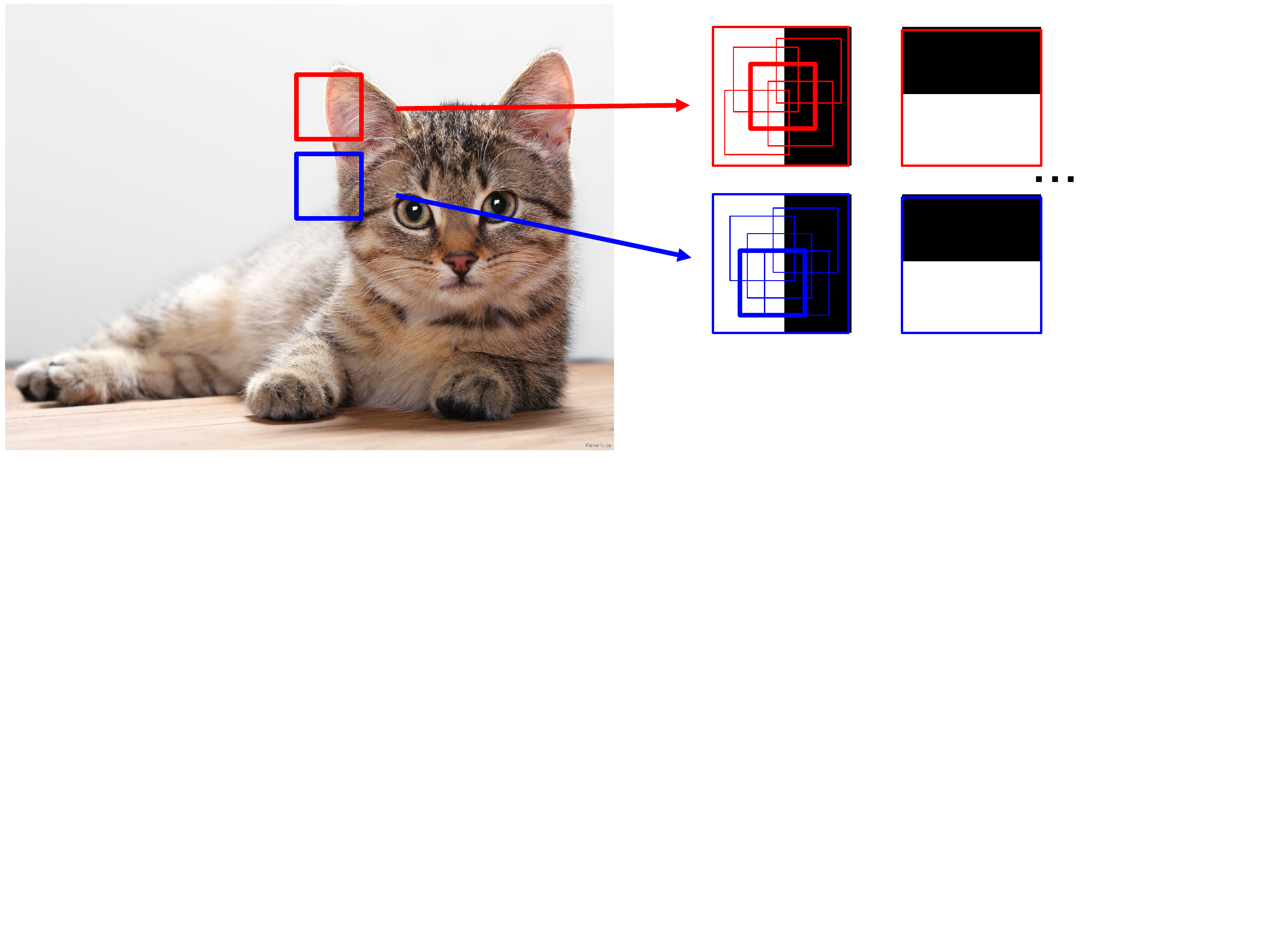}\\
    (a)&(b)
  \end{tabular}
  \caption{(a) Standard max-pooled convolution: For each filter we look for
    its best match within a small window in the data layer. (b) Proposed
    epitomic convolution (mini-epitome variant): For input data patches
    sparsely sampled on a regular grid we look for their best match in each
    mini-epitome.}
  \label{fig:epitome_diagram}
\end{figure}

\subsection{Mini-Epitomic deep networks}

We first describe a single layer of the mini-epitome variant of the proposed
model, with reference to \fig~\ref{fig:epitome_diagram}. In standard
max-pooled convolution, we have a dictionary of $K$ filters of spatial size
$\by{W}{W}$ pixels spanning $C$ channels, which we represent as real-valued
vectors $\{\wv_k\}_{k=1}^K$ with $W \cdot W \cdot C$ elements. We apply each
of them in a convolutional fashion to every $\by{W}{W}$ input patch
$\{\xv_i\}$ densely extracted from each position in the input layer which also
has $C$ channels. A reduced resolution output map is produced by computing the
maximum response within a small $\by{D}{D}$ window of displacements $p \in
\mathcal{N}_{input}$ around positions $i$ in the input map which are $D$
pixels apart from each other. 
% The sub-sampling stride is typically equal to the pooling window size $D$.
The output map $\{z_{i,k}\}$ of standard max-pooled convolution has spatial
resolution reduced by a factor of $D$ across each dimension and will consist
of $K$ channels, one for each of the $K$ filters. Specifically:
\begin{equation}
  (z_{i,k},p_{i,k}) \leftarrow \max_{p \in \mathcal{N}_{image}}
  \xv_{i+p}^T \wv_k \,
  \label{eq:conv-max}
\end{equation}
where $p_{i,k}$ points to the input layer position where the maximum is
attained.

In the proposed epitomic convolution scheme we replace the filters with larger
mini-epitomes $\{\vv_k\}_{k=1}^K$ of spatial size $\by{V}{V}$ pixels, where $V
= W+D-1$. Each mini-epitome contains $D^2$ filters $\{\wv_{k,p}\}_{k=1}^K$ of
size $\by{W}{W}$, one for each of the $\by{D}{D}$ displacements $p \in
\mathcal{N}_{epit}$ in the epitome. We \emph{sparsely} extract patches
$\{\xv_i\}$ from the input layer on a regular grid with stride $D$ pixels. In
the proposed epitomic convolution model we reverse the role of filters and
input layer patches, computing the maximum response over epitomic positions
rather than input layer positions:
\begin{equation}
  (y_{i,k},p_{i,k}) \leftarrow \max_{p \in \mathcal{N}_{epitome}}
  \xv_i^T \wv_{k,p} \,
  \label{eq:epitomic-conv}
\end{equation}
where $p_{i,k}$ now points to the position in the epitome where the maximum is
attained. Since the input position is fixed, we can think of epitomic matching
as an input-centered dual alternative to the filter-centered standard max-pooling.

Computing the maximum response over filters rather than image
positions resembles the maxout scheme of \cite{GWMCB13}, yet in the proposed
model the filters within the epitome are constrained to share values in their
area of overlap.

Similarly to max-pooled convolution, the epitomic convolution output map
$\{y_{i,k}\}$ has $K$ channels and is subsampled by a factor of $D$ across
each spatial dimension. Epitomic convolution has the same computational cost
as max-pooled convolution. For each output map value, they both require
computing $D^2$ inner products followed by finding the maximum
response. Epitomic convolution requires $D^2$ times more work per input patch,
but this is fully offset by the fact that we extract input patches sparsely
with a stride of $D$ pixels.

Similarly to standard max-pooling, the main computational primitive is
multi-channel convolution with the set of filters in the epitomic dictionary,
which we implement as matrix-matrix multiplication and carry out on the GPU,
using the cuBLAS library.

To build a deep epitomic model, we stack multiple epitomic convolution layers
on top of each other. The output of each layer passes through a rectified
linear activation unit $y_{i,k} \leftarrow \max(y_{i,k} + \beta_k, 0)$ and fed
as input to the subsequent layer, where $\beta_k$ is the bias. Similarly to
\cite{KSH13}, the final two layers of our network for Imagenet image
classification are fully connected and are regularized by dropout. We learn
the model parameters (epitomic weights and biases for each layer) in a
supervised fashion by error back propagation. We present full details of our
model architecture and training methodology in the experimental section.

\begin{figure*}[!tbp]
  \centering
% figs/visualize/alexnet2/conv1_0.eps   figs/visualize/epitonw/conv1_2.eps       figs/visualize/toponw/conv1_0.eps
% figs/visualize/epito2b30/conv1_0.eps  figs/visualize/overfeat2b30/conv1_0.eps  figs/visualize/toponw/conv1_2.eps
% figs/visualize/epitonb30/conv1_0.eps  figs/visualize/overfeatnb30/conv1_0.eps
% figs/visualize/epitonw/conv1_0.eps    figs/visualize/topon2b30/conv1_0.eps
  \begin{tabular}{c c c c c c}
    \multicolumn{3}{c}{\includegraphics[scale=1.4]{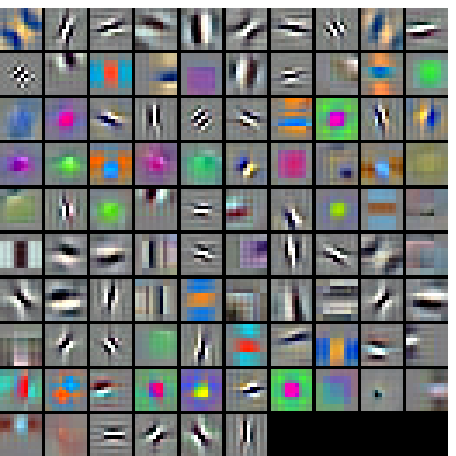}}&
    \multicolumn{3}{c}{\includegraphics[scale=1.4]{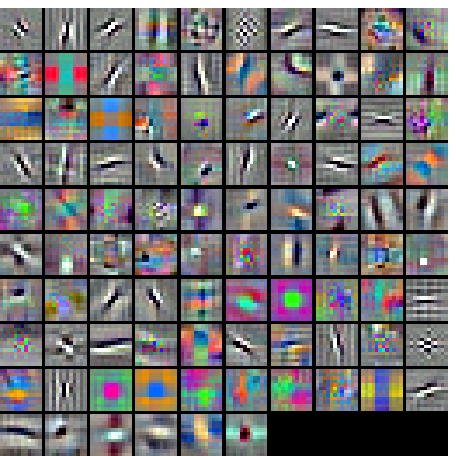}}\\
    \multicolumn{3}{c}{(a) Mini-epitomes}&
    \multicolumn{3}{c}{(b) Mini-epitomes + normalization}\\
    \\
    \multicolumn{2}{c}{\includegraphics[scale=1.4]{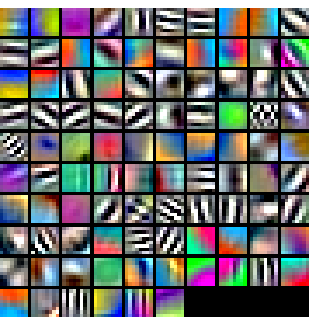}}&
    \multicolumn{2}{c}{\includegraphics[scale=1.4]{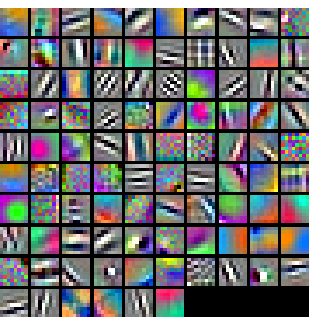}}&
    \multicolumn{2}{c}{\includegraphics[scale=1.4]{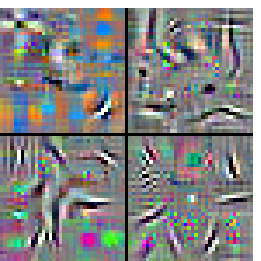}}\\
    \multicolumn{2}{c}{(c) Max-pooling}&
    \multicolumn{2}{c}{(d) Max-pooling + normalization}&
    \multicolumn{2}{c}{(e) Topographic + normaliz.}\\
  \end{tabular}
  \caption{Filters at the first convolutional layer for different models
    trained on Imagenet, shown at the same scale. For all models the input
    color image patch has size \by{8}{8} pixels. (a) Proposed
    \textsl{Epitomic} model with 96 mini-epitomes, each having size
    \by{12}{12} pixels. (b) Same as (a) with mean+contrast normalization. (c)
    Baseline \textsl{Max-Pool} model with 96 filters of size \by{8}{8} pixels
    each. (d) Same as (c) with mean+contrast normalization. (e) Proposed
    \textsl{Topographic} model with 4 epitomes of size \by{36}{36} pixels each
    and mean+contrast normalization.}
  \label{fig:visualize_conv1}
\end{figure*}

\subsection{Topographic deep networks}

We have also experimented with a topographic variant of the proposed deep
epitomic network. For this we use a dictionary with just a few large epitomes
of spatial size $\by{V}{V}$ pixels, with $V \ge W+D-1$. We retain the local
maximum responses over $\by{D}{D}$ neighborhoods spaced $D$ pixels apart in
each of the epitomes, thus yielding $( \lfloor ((V-W+1)-D)/D \rfloor + 1)^2$
output values for each of the $K$ epitomes in the dictionary. The
mini-epitomic variant can be considered as a special case of the topographic
one when $V = W+D-1$.

\subsection{Optional mean and contrast normalization}
\label{sec:mean_con_norm}

Motivated by \cite{ZeFe13b}, we have also explored the effect of filter mean
and contrast normalization on deep epitomic network training. More
specifically, we considered a variant of the model where the epitomic
convolution responses are computed as:
\begin{equation}
  (y_{i,k},p_{i,k}) \leftarrow \max_{p \in \mathcal{N}_{epitome}}
  \frac{\xv_i^T \bar{\wv}_{k,p}}{\norm{\bar{\wv}_{k,p}}_\lambda} \,
  \label{eq:epitomic-conv-norm}
\end{equation}
where $\bar{\wv}_{k,p}$ is a mean-normalized version of the filters and
$\norm{\bar{\wv}_{k,p}}_\lambda \triangleq (\bar{\wv}_{k,p}^T \bar{\wv}_{k,p}
+ \lambda)^{1/2}$ is their contrast, with $\lambda = 0.01$ a small positive
constant. This normalization requires only a slight modification of the
stochastic gradient descent update formula and incurs negligible computational
overhead. Note that the contrast normalization explored here is slightly
different than the one in \cite{ZeFe13b}, who only scale down the filters
whenever their contrast exceeds a pre-defined threshold.

We have found the mean and contrast normalization of
Eq.~\eqref{eq:epitomic-conv-norm} to be crucial for learning the topographic
version of the proposed model. We have also found that it significantly
accelerates learning of the mini-epitome version of the proposed model, as
well as the standard max-pooled convolutional model, without however
significantly affecting the final performance of these two model.

\section{Image Classification Experiments}

\subsection{Image classification tasks}

We have performed most of our experimental investigation on the Imagenet
ILSVRC-2012 dataset \cite{DDSL+09}, focusing on the task of image
classification. This dataset contains more than 1.2 million training images,
50,000 validation images, and 100,000 test images. Each image is assigned to
one out of 1,000 possible object categories. Performance is evaluated using
the top-5 classification error. Such large-scale image datasets have proven so
far essential to successfully train big deep neural networks with supervised
criteria.

Similarly to other recent works \cite{ZeFe13b, RASC14, CSVZ14}, we also
evaluate deep epitomic networks trained on Imagenet as a black-box visual
feature front-end on the Caltech-101 benchmark \cite{FFP04}. This involves
classifying images into one out of 102 possible image classes. We further
consider two standard classification benchmarks involving thumbnail-sized
images, the MNIST digit \cite{LeCo98} and the CIFAR-10 \cite{Kriz09}, both
involving classification into 10 possible classes.

\subsection{Network architecture and training methodology}

For our Imagenet experiments, we compare the proposed deep mini-epitomic and
topographic deep networks with deep convolutional networks employing standard
max-pooling. For fair comparison, we use as similar architectures as possible,
involving in all cases six convolutional layers, followed by two
fully-connected layers and a 1000-way softmax layer. We use rectified linear
activation units throughout the network. Similarly to \cite{KSH13}, we apply
local response normalization (LRN) to the output of the first two
convolutional layers and dropout to the output of the two fully-connected
layers.

\begin{table}[t]
\setlength{\tabcolsep}{3pt}
\begin{center}
\scalebox{1.00} {
\begin{tabular}{|l||c|c|c|c|c|c||c|c||c|}
  \hline  
  Layer           &   1 & 2 & 3 & 4 & 5 & 6 & 7 & 8   & Out \\
  \hline
  \hline
  Type            & conv +   & conv +   & conv & conv & conv & conv + & full + & full + & full \\
                  & lrn + max& lrn + max&      &      &      & max    & dropout& dropout&      \\ \hline
  Output channels &  96      & 192      & 256  & 384  & 512  & 512    & 4096   &  4096  & 1000 \\ \hline
  Filter size     &  8x8     & 6x6      & 3x3  & 3x3  & 3x3  & 3x3    & -    & -    & - \\ \hline
  Input stride    &  2x2     & 1x1      & 1x1  & 1x1  & 1x1  & 1x1    & -    & -    & - \\ \hline
  Pooling size    &  3x3     & 2x2      & -    & -    & -    & 3x3    & -    & -    & - \\ \hline
\end{tabular}
}
\caption{Architecture of the baseline \textsl{Max-Pool} convolutional network.}
\label{tab:max_pool_net}
\end{center}
\end{table}

The architecture of our baseline \textsl{Max-Pool} network is specified on
Table~\ref{tab:max_pool_net}. It employs max-pooling in the convolutional
layers 1, 2, and 6. To accelerate computation, it uses an image stride equal
to 2 pixels in the first layer. It has a similar structure with the Overfeat
model \cite{SEZM+14}, yet significantly fewer neurons in the convolutional
layers 2 to 6. Another difference with \cite{SEZM+14} is the use of LRN, which
to our experience facilitates training.

The architecture of the proposed \textsl{Epitomic} network is specified on
Table~\ref{tab:mini_epitome_net}. It has exactly the same number of neurons at
each layer as the \textsl{Max-Pool} model but it uses mini-epitomes in place
of convolution + max pooling at layers 1, 2, and 6. It uses the same filter
sizes with the \textsl{Max-Pool} model and the mini-epitome sizes have been
selected so as to allow the same extent of translation invariance as the
corresponding layers in the baseline model. We use input image stride equal to
4 pixels and further perform epitomic search with stride equal to 2 pixels in
the first layer to also accelerate computation.

\begin{table}[t]
\setlength{\tabcolsep}{3pt}
\begin{center}
\scalebox{1.00} {
\begin{tabular}{|l||c|c|c|c|c|c||c|c||c|}
  \hline  
  Layer           &   1 & 2 & 3 & 4 & 5 & 6 & 7 & 8   & Out \\
  \hline
  \hline
  Type            & epit-conv& epit-conv& conv & conv & conv & epit-conv & full + & full + & full \\
                  & + lrn    & + lrn    &      &      &      &           & dropout& dropout&      \\ \hline
  Output channels &  96      & 192      & 256  & 384  & 512  & 512       & 4096   &  4096  & 1000 \\ \hline
  Epitome size    &  12x12   & 8x8      &  -   &  -   &  -   & 5x5       & -    & -    & - \\ \hline
  Filter size     &  8x8     & 6x6      & 3x3  & 3x3  & 3x3  & 3x3       & -    & -    & - \\ \hline
  Input stride    &  4x4     & 3x3      & 1x1  & 1x1  & 1x1  & 3x3       & -    & -    & - \\ \hline
  Epitome stride  &  2x2     & 1x1      &  -   &  -   &  -   & 1x1       & -    & -    & - \\ \hline
\end{tabular}
}
\caption{Architecture of the proposed \textsl{Epitomic} convolutional network.}
\label{tab:mini_epitome_net}
\end{center}
\end{table}

The architecture of our second proposed \textsl{Topographic} network is
specified on Table~\ref{tab:topographic_net}. It uses four epitomes at layers 1,
2 and eight epitomes at layer 6 to learn topographic feature maps. It uses the
same filter sizes as the previous two models and the epitome sizes have been
selected so as each layer produces roughly the same number of output channels
when allowing the same extent of translation invariance as the corresponding
layers in the other two models.

\begin{table}[t]
\setlength{\tabcolsep}{3pt}
\begin{center}
\scalebox{1.00} {
\begin{tabular}{|l||c|c|c|c|c|c||c|c||c|}
  \hline  
  Layer           &   1 & 2 & 3 & 4 & 5 & 6 & 7 & 8   & Out \\
  \hline
  \hline
  Type            & epit-conv& epit-conv& conv & conv & conv & epit-conv & full + & full + & full \\
                  & + lrn    & + lrn    &      &      &      &           & dropout& dropout&      \\ \hline
  Output channels &  4x25    & 4x49     & 256  & 384  & 512  & 8x64      & 4096   &  4096  & 1000 \\ \hline
  Epitome size    &  36x36   & 26x26    &  -   &  -   &  -   & 26x26     & -    & -    & - \\ \hline
  Filter size     &  8x8     & 6x6      & 3x3  & 3x3  & 3x3  & 3x3       & -    & -    & - \\ \hline
  Input stride    &  4x4     & 3x3      & 1x1  & 1x1  & 1x1  & 3x3       & -    & -    & - \\ \hline
  Epitome stride  &  2x2     & 1x1      &  -   &  -   &  -   & 1x1       & -    & -    & - \\ \hline
  Epit. pool size &  3x3     & 3x3      & -    & -    & -    & 3x3       & -    & -    & - \\ \hline
\end{tabular}
}
\caption{Architecture of the proposed \textsl{Topographic} convolutional network.}
\label{tab:topographic_net}
\end{center}
\end{table}

We have also tried variants of the three models above where we activate the
mean and contrast normalization scheme of Section~\ref{sec:mean_con_norm} in
layers 1, 2, and 6 of the network.

We followed the methodology of \cite{KSH13} in training our models. We used
stochastic gradient ascent with learning rate initialized to 0.01 and
decreased by a factor of 10 each time the validation error stopped
improving. We used momentum equal to 0.9 and mini-batches of 128 images. The
weight decay factor was equal to $\by{5}{10^{-4}}$. Importantly, weight decay
needs to be turned off for the layers that use mean and contrast
normalization. Training each of the three models takes two weeks using a
single NVIDIA Titan GPU. Similarly to \cite{CSVZ14}, we resized the training
images to have small dimension equal to 256 pixels while preserving their
aspect ratio and not cropping their large dimension. We also subtracted for
each image pixel the global mean RGB color values computed over the whole
Imagenet training set. During training, we presented the networks with
$\by{220}{220}$ crops randomly sampled from the resized image area, flipped
left-to-right with probability 0.5, also injecting global color noise exactly
as in \cite{KSH13}. During evaluation, we presented the networks with 10
regularly sampled image crops (center + 4 corners, as well as their
left-to-right flipped versions).

\subsection{Weight visualization}

We visualize in Figure~\ref{fig:visualize_conv1} the layer weights at the first
layer of the networks above. The networks learn receptive fields sensitive to
edge, blob, texture, and color patterns. 

\subsection{Classification results}

We report at Table~\ref{tab:imagenet_results} our results on the Imagenet
ILSVRC-2012 benchmark, also including results previously reported in the
literature \cite{KSH13, ZeFe13b, SEZM+14}. These all refer to the top-5 error
on the validation set and are obtained with a single network. Our best result
at 13.6\% with the proposed \textsl{Epitomic-Norm} network is 0.6\% better
than the baseline \textsl{Max-Pool} result at 14.2\% error. Our
\textsl{Topographic-Norm} network scores less well, yielding 15.4\% error
rate, which however is still better than \cite{KSH13, ZeFe13b}. Mean and
contrast normalization had little effect on final performance for the
\textsl{Max-Pool} and \textsl{Epitomic} models, but we found it essential for
learning the \textsl{Topographic} model. The improved performance that we got
with the \textsl{Max-Pool} baseline network compared to Overfeat
\cite{SEZM+14} is most likely due to our use of LRN and aspect ratio
preserving image resizing. When preparing this manuscript, we became aware of
the work of \cite{CSVZ14} that reports an even lower 13.1\% error rate with a
max-pooled network, using however significantly more neurons than we do in the
convolutional layers 2 to 5.

\begin{table}[t]
\setlength{\tabcolsep}{3pt}
\begin{center}
\scalebox{0.9} {
\begin{tabular}{|l||c|c|c||c|c|c|c|c|}
  \hline  
  Model       & Krizhevsky  & Zeiler-Fergus  & Overfeat   & Max-Pool &  Max-Pool & Epitomic  & Epitomic  & Topographic \\
              & \cite{KSH13}& \cite{ZeFe13b} &\cite{SEZM+14}&        &  + norm   &           & + norm    &  + norm     \\ \hline
  Top-5 Error &   18.2\%    &    16.0\%      &  14.7\%    &  14.2\%  &  14.4\%   &\bf{13.7\%}&\bf{13.6\%}&   15.4\%    \\
  \hline
\end{tabular}
}
\caption{Imagenet ILSVRC-2012 top-5 error on validation set. All performance
  figures are obtained with a single network, averaging classification
  probabilities over 10 image crops (center + 4 corners, as well as their
  left-to-right flipped versions).}
\label{tab:imagenet_results}
\end{center}
\end{table}

We next assess the quality of the proposed model trained on Imagenet as
black-box feature extractor for Caltech-101 image classification. For this
purpose, we used the 4096-dimensional output of the last fully-connected
layer, without doing any fine-tuning of the network weights for the new
task. We trained a 102-way SVM classifier using \textsl{libsvm} \cite{ChLi11}
and the default regularization parameter. For this experiment we just resized
the Caltech-101 images to size \by{220}{220} without preserving their aspect
ratio and computed a single feature vector per image. We normalized the
feature vector to have unit length before feeding it into the SVM. We report
at Table~\ref{tab:caltech101_results} the mean classification accuracy
obtained with the different networks. The proposed \textsl{Epitomic} model
performs at 87.8\%, 0.5\% better than the baseline \textsl{Max-Pool} model. 

% Caltech-101 results (center crop only)
% alexnet2: Accuracy = 88.215% (5367/6084) / Mean accuracy = 84.52
% epitonb30: Accuracy = 89.3984% (5439/6084) / Mean accuracy = 87.37
% epito2b30: Accuracy = 89.5464% (5448/6084) / Mean accuracy = 87.85
% overfeat2b30: Accuracy = 89.119% (5422/6084) / Mean accuracy = 87.31
% overfeatnb30: Accuracy = 87.426% (5319/6084) / Mean accuracy = 85.34
% topon2b30: Accuracy = 87.1959% (5305/6084) / Mean accuracy = 85.76
\begin{table}[t]
\setlength{\tabcolsep}{3pt}
\begin{center}
\scalebox{0.9} {
\begin{tabular}{|l||c||c|c|c|c|c|}
  \hline  
  Model       & Zeiler-Fergus  &  Max-Pool &  Max-Pool & Epitomic  & Epitomic  & Topographic \\
              & \cite{ZeFe13b} &           &  + norm   &           & + norm    &  + norm     \\ \hline
  Mean Accuracy &    86.5\%    &   87.3\%  &  85.3\%   &\bf{87.8\%}&  87.4\%   &   85.8\%    \\
  \hline
\end{tabular}
}
\caption{Caltech-101 mean accuracy with deep networks pretrained on Imagenet.}
\label{tab:caltech101_results}
\end{center}
\end{table}

We have also performed experiments with the epitomic model on classifying
small images on the MNIST and CIFAR-10 datasets. For these tasks we have
trained much smaller networks from scratch, using three epitomic convolutional
layers, followed by one fully-connected layer and the final softmax
classification layer. Because of the small training set sizes, we have found
it beneficial to also employ dropout regularization in the epitomic
convolution layers. At Table~\ref{tab:mnist_cifar10_results} we report the
classification error rates we obtained. Our results are comparable to maxout
\cite{GWMCB13}, which achieves state-of-art results on these tasks.

\begin{table}[t]
\begin{center}
\scalebox{0.9} {
\begin{tabular}{c c}
  \begin{tabular}{|l||c|c|}
    \hline  
    Model       & Maxout        & Epitomic \\ \hline
    Error rate  & 0.45\%        & 0.44\%   \\ \hline
  \end{tabular}
  &
  \begin{tabular}{|l||c|c|}
    \hline  
    Model       & Maxout        & Epitomic \\ \hline
    Error rate  & 9.38\%        & 9.43\%   \\ \hline
  \end{tabular}
  \\
  (a) MNIST & (b) CIFAR-10
\end{tabular}
}
\caption{Classification error rates on small image datasets for maxout
  \cite{GWMCB13} and the proposed mini-epitomic deep network: (a) MNIST. (b)
  CIFAR-10.}
\label{tab:mnist_cifar10_results}
\end{center}
\end{table}

\subsection{Mean-contrast normalization and convergence speed}

\begin{figure}[!tbp]
  \centering
  \includegraphics[width=0.8\columnwidth]{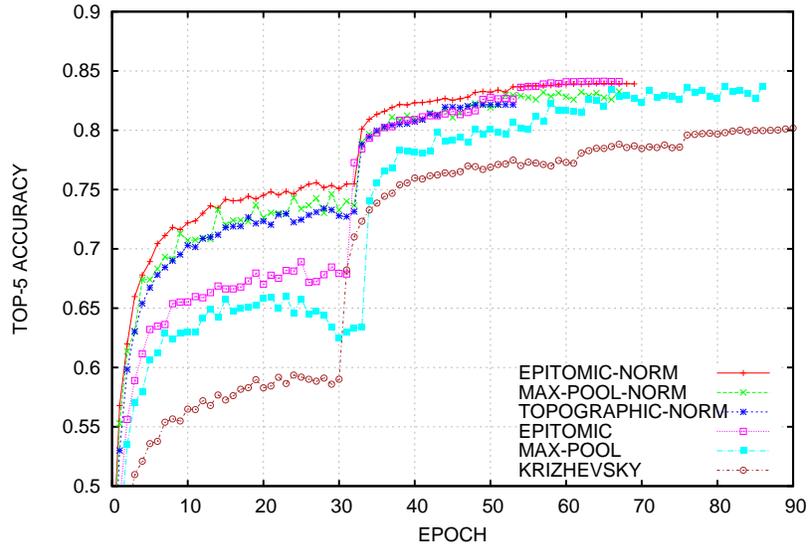}
  \caption{Top-5 validation set accuracy (center non-flipped crop only) for
    different models and normalization.}
  \label{fig:imagenet_optim}
\end{figure}

We comment on the learning speed and convergence properties of the different
models we experimented with on Imagenet. We show in
Figure~\ref{fig:imagenet_optim} how the top-5 validation error improves as
learning progresses for the different models we tested, with or without
mean+contrast normalization. For reference, we also include a corresponding
plot we re-produced for the original model of Krizhevsky \etal
\cite{KSH13}. We observe that mean+contrast normalization significantly
accelerates convergence of both epitomic and max-pooled models, without
however significantly influencing the final model quality. The epitomic models
exhibit somewhat improved convergence behavior during learning compared to the
max-pooled baselines whose performance fluctuates more.

\section{Conclusions}

In this paper we have explored the potential of the epitomic representation as
a building block for deep neural networks. We have shown that an epitomic
layer can successfully substitute a pair of consecutive convolution and
max-pooling layers. We have proposed two deep epitomic variants, one featuring
mini-epitomes that empirically performs best in image classification, and one
featuring large epitomes and learns topographically organized feature maps. We
have shown that the proposed epitomic model performs around 0.5\% better than
the max-pooled baseline on the challenging Imagenet benchmark and other image
classification tasks. 

In future work, we are very interested in developing methods for unsupervised
or semi-supervised training of deep epitomic models, exploiting the fact that
the epitomic representation is more amenable than max-pooling for
incorporating image reconstruction objectives.

\paragraph{Reproducibility} We implemented the proposed methods by extending
the excellent Caffe software framework \cite{Jia13}. When this work gets
published we will publicly share our source code and configuration files with
exact parameters fully reproducing the results reported in this paper.

\paragraph{Acknowledgments} We gratefully acknowledge the support of NVIDIA
Corporation with the donation of GPUs used for this research.

%{\small

\bibliographystyle{ieee}
\bibliography{IEEEabrv,biblio_preamble_abrv,biblio_gpapan}

\newcommand{\noopsort}[1]{} \newcommand{\printfirst}[2]{#1}
  \newcommand{\singleletter}[1]{#1} \newcommand{\switchargs}[2]{#2#1}
\begin{thebibliography}{10}\itemsep=-1pt

\bibitem{AhEl08}
M.~Aharon and M.~Elad.
\newblock Sparse and redundant modeling of image content using an
  image-signature-dictionary.
\newblock {\em {SIAM} J. Imaging Sci.}, 1(3):228--247, 2008.

\bibitem{BMBP11}
L.~Beno{\^\i}t, J.~Mairal, F.~Bach, and J.~Ponce.
\newblock Sparse image representation with epitomes.
\newblock In {\em Proc. {CVPR}}, pages 2913--2920, 2011.

\bibitem{ChLi11}
C.-C. Chang and C.-J. Lin.
\newblock {LIBSVM}: a library for support vector machines.
\newblock {\em {ACM} Trans. on Intel. Systems and Tech.}, 2(3), 2011.

\bibitem{CSVZ14}
K.~Chatfield, K.~Simonyan, A.~Vedaldi, and A.~Zisserman.
\newblock Return of the devil in the details: Delving deep into convolutional
  nets.
\newblock arXiv, 2014.

\bibitem{DDSL+09}
J.~Deng, W.~Dong, R.~Socher, L.~Li-Jia, K.~Li, and L.~Fei-Fei.
\newblock Imagenet: A large-scale hierarchical image database.
\newblock In {\em Proc. {CVPR}}, 2009.

\bibitem{FFP04}
L.~Fei-Fei, R.~Fergus, and P.~Perona.
\newblock Learning generative visual models from few training examples: An
  incremental {B}ayesian approach tested on 101 object categories.
\newblock In {\em Proc. {CVPR} Workshop}, 2004.

\bibitem{GDDM14}
R.~Girshick, J.~Donahue, T.~Darrell, and J.~Malik.
\newblock Rich feature hierarchies for accurate object detection and semantic
  segmentation.
\newblock In {\em Proc. {CVPR}}, 2014.

\bibitem{GWMCB13}
I.~Goodfellow, D.~Warde-Farley, M.~Mirza, A.~Courville, and Y.~Bengio.
\newblock Maxout networks.
\newblock In {\em Proc. {ICML}}, 2013.

\bibitem{HyHo01}
A.~Hyv{\"a}rinen, P.~Hoyer, and M.~Inki.
\newblock Topographic independent component analysis.
\newblock {\em Neur. Comp.}, 13(7):1527--1558, 2001.

\bibitem{JKRL09}
K.~Jarrett, K.~Kavukcuoglu, M.~Ranzato, and Y.~LeCun.
\newblock What is the best multi-stage architecture for object recognition?
\newblock In {\em Proc. {ICCV}}, pages 2146--2153, 2009.

\bibitem{Jia13}
Y.~Jia.
\newblock {Caffe}: An open source convolutional architecture for fast feature
  embedding.
\newblock \url{http://caffe.berkeleyvision.org/}, 2013.

\bibitem{JFK03}
N.~Jojic, B.~Frey, and A.~Kannan.
\newblock Epitomic analysis of appearance and shape.
\newblock In {\em Proc. {ICCV}}, pages 34--41, 2003.

\bibitem{KRFL09}
K.~Kavukcuoglu, M.~Ranzato, R.~Fergus, and Y.~LeCun.
\newblock Learning invariant features through topographic filter maps.
\newblock In {\em Proc. {CVPR}}, 2009.

\bibitem{Kriz09}
A.~Krizhevsky.
\newblock Learning multiple layers of features from tiny images.
\newblock Technical report, 2009.

\bibitem{KSH13}
A.~Krizhevsky, I.~Sutskever, and G.~Hinton.
\newblock {ImageNet} classification with deep convolutional neural networks.
\newblock In {\em Proc. {NIPS}}, 2013.

\bibitem{LRMD+12}
Q.~Le, M.~Ranzato, R.~Monga, M.~Devin, G.~Corrado, K.~Chen, J.~Dean, and A.~Ng.
\newblock Building high-level features using large scale unsupervised learning.
\newblock In {\em Proc. {ICML}}, 2012.

\bibitem{LBBH98}
Y.~LeCun, L.~Bottou, Y.~Bengio, and P.~Haffner.
\newblock Gradient-based learning applied to document recognition.
\newblock {\em Proc. {IEEE}}, 86(11):2278--2324, 1998.

\bibitem{LeCo98}
Y.~LeCun and C.~Cortes.
\newblock The {MNIST} database of handwritten digits, 1998.

\bibitem{LGRN09}
H.~Lee, R.~Grosse, R.~Ranganath, and A.~Y. Ng.
\newblock Convolutional deep belief networks for scalable unsupervised learning
  of hierarchical representations.
\newblock In {\em Proc. {ICML}}, 2009.

\bibitem{Lowe04}
D.~Lowe.
\newblock Distinctive image features from scale-invariant keypoints.
\newblock {\em {IJCV}}, 60(2):91--110, 2004.

\bibitem{OWH06}
S.~Osindero, M.~Welling, and G.~Hinton.
\newblock Topographic product models applied to natural scene statistics.
\newblock {\em Neur. Comp.}, 18:381--414, 2006.

\bibitem{OuWa13}
W.~Ouyang and X.~Wang.
\newblock Joint deep learning for pedestrian detection.
\newblock In {\em Proc. {ICCV}}, 2013.

\bibitem{PCY14}
G.~Papandreou, L.-C. Chen, and A.~Yuille.
\newblock Modeling image patches with a generic dictionary of mini-epitomes.
\newblock In {\em Proc. {CVPR}}, 2014.

\bibitem{RASC14}
A.~Razavian, H.~Azizpour, J.~Sullivan, and S.~Carlsson.
\newblock {CNN} features off-the-shelf: An astounding baseline for recognition.
\newblock In {\em Proc. {CVPR} Workshop}, 2014.

\bibitem{RiPo99}
M.~Riesenhuber and T.~Poggio.
\newblock Hierarchical models of object recognition in cortex.
\newblock {\em Nature neuroscience}, 2(11):1019--1025, 1999.

\bibitem{SEZM+14}
P.~Sermanet, D.~Eigen, X.~Zhang, M.~Mathieu, R.~Fergus, and Y.~LeCun.
\newblock Overfeat: Integrated recognition, localization and detection using
  convolutional networks.
\newblock 2014.

\bibitem{ZeFe13a}
M.~Zeiler and R.~Fergus.
\newblock Stochastic pooling for regularization of deep convolutional neural
  networks.
\newblock 2013.

\bibitem{ZeFe13b}
M.~Zeiler and R.~Fergus.
\newblock Visualizing and understanding convolutional networks.
\newblock arXiv, 2013.

\bibitem{ZKTF10}
M.~Zeiler, D.~Krishnan, G.~Taylor, and R.~Fergus.
\newblock Deconvolutional networks.
\newblock In {\em Proc. {CVPR}}, 2010.

\end{thebibliography}

%}

\end{document}